# RADIAL BASIS FUNCTION PROCESS NEURAL NETWORK TRAINING BASED ON GENERALIZED FRÉCHET DISTANCE AND GA-SA HYBRID STRATEGY

Wang Bing[1], Meng Yao-hua[2], Yu Xiao-hong[3]

[1]School of Computer & Information Technology, Northeast Petroleum University, Daqing , China
[2] Communication Research Center, Harbin Institute of Technology, Harbin , China
[3] Exploration and Development Research Institute, Daqing Oilfield Company, Daqing, China

**ABSTRACT**

*For learning problem of Radial Basis Function Process Neural Network (RBF-PNN), an optimization training method based on GA combined with SA is proposed in this paper. Through building generalized Fréchet distance to measure similarity between time-varying function samples, the learning problem of radial basis centre functions and connection weights is converted into the training on corresponding discrete sequence coefficients. Network training objective function is constructed according to the least square error criterion, and global optimization solving of network parameters is implemented in feasible solution space by use of global optimization feature of GA and probabilistic jumping property of SA . The experiment results illustrate that the training algorithm improves the network training efficiency and stability.*

**KEYWORDS**

*Radial Basis Function Process Neural Network, Training Algorithm,* Generalized Fréchet Distance*, GA-SA Hybrid Optimization*

## 1. INTRODUCTION

Radial basis function neural network (RBFNN) is a three-layer feedforward neural network model and was put forward by Powell, Broomhead and Lowe[1,2] in the mid-1980s. It implements nonlinear mapping by change property parameters of a neuron and improves learning rate by linearization of connection weight adjustment. RBFNN has been applied in many fields such as function approximation, spline interpolation and pattern recognition [3,4]. However, in practical engineering, many systems' inputs are dependent on time; the inputs of existing RBFNN model are usually constant. They have nothing to do with the time, that is, the input/output is instantaneous relationship between geometric points. Its information processing mechanism cannot directly reflect the characteristics of time-varying process input signals and the dynamic effect relationship between variables. Therefore, a radial basis function process neural network (RBF-PNN) is established in literature [5], and its inputs can be time-varying process signal directly. Through space-time aggregation of kernel function neurons to find sample pattern process characteristics, the RBF-PNN has been applied in dynamic sample clustering and time-varying signal identification effectively.

In practice, time-varying sample data is usually discrete time series obtained by sampling, the existing RBF-PNN is unable to deal with discrete input directly. It need to fit discrete samples, thus fitting error exists. In subsequent orthogonal basis expansion, the finite terms of basis function also bring truncation error. In addition, learning efficiency of traditional RBF-PNN algorithm based on gradient descent combined with basis expansion is low, and at the same time it affects stability and generalization ability of the network that there are greater freedom constraints among network parameters.

In recent years, evolutionary computation-based optimization technology is widely used to solve optimal solution in various engineering problems, and has been effectively applied in complex function calculation, process optimization, system identification and control, and so on [6-8]. Genetic algorithm (GA) [9] is a bionic global optimization method that simulates Darwin's natural selection and natural biological evolution process. It is simple, general, robust, suitable for parallel processing, as well as efficient and practical, and has been widely applied in various fields. However, GA has some shortcomings, e.g. immature convergence and slow iterative rate. If probabilistic jumping property of simulated annealing (SA) is made use of, the GA combined with SA are applied into the training of RBF-PNN, this will improve RBF-PNN learning property greatly.

Fréchet distance [10] is a judgment to measure the similarity between polygonal curves. In applications, dynamic samples normally consist of a sequence of discrete sampling points, therefore, Eiter and Mannila [11] put forward discrete Fréchet distance on the basis of continuous Fréchet distance, and it achieved good application effect in protein structure prediction[12], online signature verification [13], etc. Time-varying process signal can be seen as a one-dimensional curve about the time; therefore, discrete Fréchet distance can be extended to time-varying function space to measure nature difference between input samples of RBF-PNN.

For the RBF-PNN training problem, we propose a training method based on generalized discrete Fréchet distance combined with GA-SA optimization in this paper. By constructing a generalized Fréchet distance norm to measure the distance between dynamic samples in time-varying function space, then the learning problem of weight functions is transformed into discrete coefficient training. According to the least square error criterion, the network training objective function is constructed, and GA-SA algorithm is adopted for global optimization solving. It is applied into EEG eye state recognition and achieves good results. RBF-PNN optimization training steps are also given in the paper.

## 2. DISTANCE MEASUREMENT BETWEEN TIME-VARYING SAMPLES BASED ON GENERALIZED FRÉCHET DISTANCE

### 2.1. Discrete Fréchet Distance

Discrete Fréchet distance is defined as follows:

(a) Given a polygonal chain $P = < p_1, p_2, \cdots, p_n >$ with $n$ vertices, a k-walk along $P$ partitions the path into $k$ disjoint non-empty subsets $\{P_i\}_{i=1,\cdots,k}$ such that $P_i = < p_{n_{i-1}+1}, \cdots, p_{n_i} >$ and $0 = n_0 < n_1 < \cdots < n_k = n$.

(b) Given two polygonal chains $A =< a_1, a_2, \cdots, a_m >$ and $B =< b_1, b_2, \cdots, b_n >$, a paired walk along $A$ and $B$ is a k-walk $\{A_i\}_{i=1,\cdots,k}$ along $A$ and a k-walk $\{B_i\}_{i=1,\cdots,k}$ along $B$ for $1 \leq i \leq k$, either $|A_i|=1$ or $|B_i|=1$ (namely, $A_i$ or $B_i$ contains exactly one vertex).

(c) The cost of a paired walk $W = \{(A_i, B_i)\}$ along $A$ and $B$ is

$$d_F^W(A,B) = \max_i \max_{(a,b) \in A_i \times B_i} dist(a,b) \quad (1)$$

The discrete Fréchet distance between two polygonal chains $A$ and $B$ is

$$d_F(A,B) = \min_W d_F^W(A,B) \quad (2)$$

The paired walk is also called the Fréchet alignment of $A$ and $B$.

## 2.2. Generalized Fréchet Distance Between Time-Varying Samples

The discrete Fréchet distance can effectively measure the difference between time-varying functions (which can be regarded as the polygonal chains), while the input of the RBF-PNN is a vector composed of time-varying functions. Thus, using the property that Euclidean distance can implement point target matching, combining the discrete Fréchet distance and Euclidean distance, a generalized Fréchet distance is established, which can measure the distance between time-varying function vector samples.

Given two time-varying function samples $\mathbf{X}(t) = (x_1(t), x_2(t), ..., x_n(t))$ and $X'(t) = (x'_1(t), x'_2(t), ..., x'_n(t))$, their generalized Fréchet distance $d(\mathbf{X}(t), X'(t))$ is defined as

$$d(\mathbf{X}(t), X'(t)) = (\sum_{i=1}^{n} d_F(x_i(t), x'_i(t))^2)^{\frac{1}{2}} \quad (3)$$

where $d_F(x_i(t), x'_i(t))$ denotes the discrete Fréchet distance between discrete sampling points corresponding to $x_i(t)$ and $x'_i(t)$.

## 3. RBF-PNN MODEL

### 3.1. Radial Basis Function Process Neuron

The structure and information transfer flow of a radial basis function process neuron are shown in Figure 1.

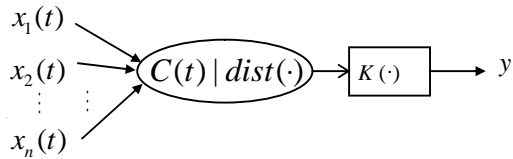

Figure 1. Radial basis function process neuron

In Figure 1, $\mathbf{X}(t) = (x_1(t), x_2(t), ..., x_n(t))$ is an input function vector; $C(t) = (c_1(t), c_2(t), ..., c_n(t))$ is a centre function vector of a radial basis function neuron; $dist(\cdot)$ is a distance measurement function between the input function vector and the centre function vector, and here we adopt the generalized Fréchet distance defined in Eq.(3). $K(\cdot)$ is a kernel function of a radial basis process neuron. It could be a Gauss function, a multiquadric function and a thin-plate spline function etc. The input/output relationship of a radial basis function process neuron is defined as:

$$y = K(dist \| X(t), C(t) \|) \qquad (4)$$

### 3.2. Radial Basis Function Process Neural Network Model

A RBF-PNN model is composed of radial basis function process neurons and other types of neurons according to certain structure relations and information transmission process. For convenience, consider a multiple input/single output feedforward network model only with one radial basis process hidden layer defined by Eq. (4). Its structure is shown in Figure 2.

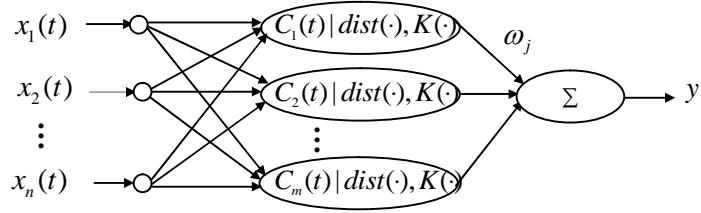

Figure 2. RBF-PNN model

In Figure 2, the input layer has $n$ neurons, the radial basis process hidden layer has $m$ neurons, and $\omega_j (j = 1, 2, ..., m)$ are the weighted coefficients to be adjusted. The input/output relationship of RBF-PNN is:

$$y = \sum_{j=1}^{m} \omega_j \cdot K(dist \| X(t), C_j(t) \|) \qquad (5)$$

### 3.3. Network Objective Function

Given $K$ learning sample functions $(X_k(t), d_k)$ where $X_k(t) = x_{k1}(t), x_{k2}(t), ..., x_{kn}(t)$, $k = 1, 2, ..., K$. Suppose that $y_k$ is a real output corresponding to the $k$ th sample input. According to the least square error criterion, the network objective function is defined as below:

$$E = \sum_{k=1}^{K} (y_k - d_k)^2 \qquad (6)$$

The kernel function of the radial basis process neuron adopts Gauss function:

$$K(d) = \exp(-d^2 / 2\sigma^2) \qquad (7)$$

where $\sigma$ is the width of Gauss function, and it needs to be determined via network training, then we have

$$E = \sum_{k=1}^{K}(y_k - d_k)^2$$
$$= \sum_{k=1}^{K}(\sum_{j=1}^{m}\omega_j \cdot \exp(-\frac{dist\|X_k(t), C_j(t)\|^2}{2\sigma_j^2}) - d_k)^2 \qquad (8)$$

In Eq.(8), since $dist\|X_k(t), C_j(t)\|$ is based on generalized discrete Fréchet distance, if discrete sampling point number of RBF-PNN input is $S$, then $\{c_{ji}^s\}_{s=1}^{S}$ is the discrete point sequence corresponding to $c_{ji}(t)$, $i=1,2,\cdots,n$, which is the $i$ th component of the $j$ th radial basis centre function vector. In conclusion, the learning objective function $E$ of RBF-PNN is a function about variable parameters $\{c_{ji}^s\}_{s=1}^{S}, \sigma_j, \omega_j$.

## 4. RBF-PNN TRAINING BASED ON GA-SA ALGORITHM

In GA [9], selection, crossover and mutation constitute its main operations. Parameter encoding, initial population, fitness function choice, genetic operation design, control parameter setting form the core content of GA. The obvious advantage of GA is to search many points in search space at the same time, and its optimal solution searching process is instructive so as to avoid the dimension disaster problem existing in some other optimization algorithms. Simulated annealing (SA)[14] algorithm is a heuristic random search algorithm. Under a certain initial temperature, accompanied by temperature fall, by use of probabilistic jumping property, the global optimal solution of the objective function is obtained in solution space. In view of the respective features of GA and SA, considering that SA can effectively prevent GA from trapping in local optimal solution, GA and SA will be combined for RBF-PNN training to obtain the optimal solution of the problem.

The training problem of RBF-PNN is to adjust parameters to minimize the objective function $E=E(\{c_{ji}^s\}_{s=1}^{S}, \sigma_j, \omega_j)$. The concrete RBF-PNN training steps based on GA-SA is described below:

**Step 1** Given initial simulated annealing temperature $t_1$, and set $k=1$.

**Step 2** Determine the population size $N$, and randomly generate initial population. Decimal number is used for chromosome encoding, and the gene number on each chromosome is the number of variables to be optimized. Set the maximum iterative number $M$ and error precision $\varepsilon$.

**Step 3** Evaluate chromosomes in the population. Since RBF-PNN training is the minimum optimization problem for the objective function, the fitness function is taken as

$$f = \frac{1}{1+E} \quad (9)$$

**Step 4** Perform genetic operations for chromosomes in the population.
(1) Selection operation. Adopt proportional selection operator, and the probability that the chromosome $X_i$ with fitness value $f_i$ is selected into the next generation is

$$p_i = f_i \bigg/ \sum_{j=1}^{M} f_j \quad (10)$$

(2) Crossover operation. For decimal number encoding, choose arithmetic crossover operator. Chromosomes $X_1$ and $X_2$ in the parent perform crossover operation with crossover probability $p_c$, the offspring chromosomes generated are

$$X_1^{'} = aX_1 + (1-a)X_2 \quad (11)$$

$$X_2^{'} = (1-a)X_1 + aX_2 \quad (12)$$

where $a$ is a parameter and $a \in (0,1)$.

(3) Mutation operation. Use uniform mutation operator. Each gene in a chromosome $X_i$ performs mutation operation with mutation probability $p_m$, namely uniformly distributed random number within the scope of gene value is used to instead of the original value with probability $p_m$.

**Step 5** Introduce the optimal retention strategy.

**Step 6** Each chromosome in the population performs simulated annealing operation.
(1) Create new gene $g^{'}(k) = g(k) + \beta$ by use of SA state generation function where $\beta \in (\frac{1}{10}\min, \frac{1}{10}\max)$ are the random disturbance and (min,max) are the scope of corresponding gene value.
(2) Calculate objective function difference value $\Delta C$ between $g^{'}(k)$ and $g(k)$.
(3) Calculate acceptance probability $p_r = \min[1, \exp(-\Delta C / t_k)]$.
(4) If $p_r > random[0,1]$ then $g(k) = g^{'}(k)$; otherwise $g(k)$ remains unchanged.
(5) Introduce the optimal retention strategy.
(6) Anneal temperature by using the annealing temperature function $t_{k+1} = vt_k$ where $v \in (0,1)$ is the annealing temperature rate, and set $k = k+1$.

**Step 7** If the objective function value corresponding to the optimal chromosome meets error requirements $\varepsilon$ or the iteration process achieves the maximum number $M$, turn to Step 8; otherwise return to Step 3.

**Step 8** Decode the optimal chromosome searched by GA and obtain the optimal parameters of RBF-PNN.

## 5. APPLICATIONS IN EEG EYE STATE CLASSIFICATION

The simulation experiments use "EEG Eye State Data Set" from UCI. The data set was collected for examining a classifier for sequential time series. The data set consists of 14 EEG values and a value indicating the eye state, and '1' indicates the eye-closed and '0' the eye-open state. All data is from one continuous EEG measurement with the Emotiv EEG Neuroheadset. The part of samples with the eye-closed state and the eye-open state are shown in Figure 3.

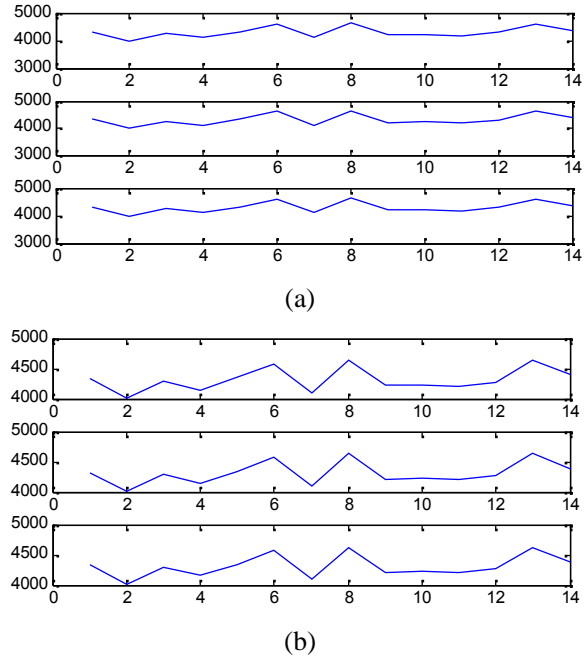

Figure 3. (a) The eye-closed state samples (b) The eye-open state samples

In the experiments, the training set includes 60 samples with each 30 samples separately from two states, and the test set consists of 60 samples. The RBF-PNN structure parameters are : 1 time-varying input node $n = 1$, 8 RBF hidden nodes $m = 8$ and 1 common output node. The discrete sampling number is $S = 14$, the population size is 25, optimization iteration number is 1000, and the error precision is 0.03. As the comparison, we also use gradient descent (GD) algorithm [5] and GA to train the RBF-PNN. Each algorithm runs 10 times, and then three algorithms are separately used to identify the test samples. The average identification accuracy is shown in Figure 4. It can be seen that the SA's join into GA prevents into local minimum effectively and improves the identification result of the networks.

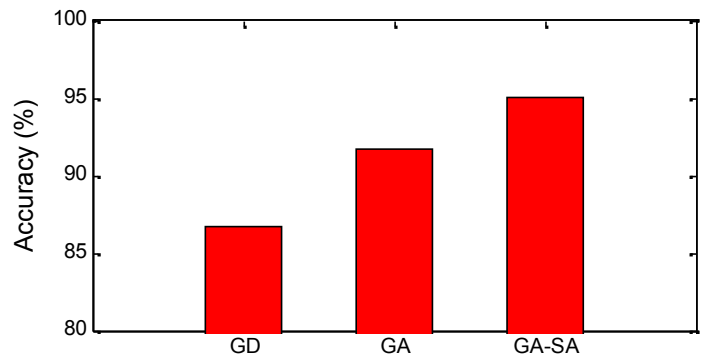

Figure 4. Accuracy of three identification algorithms

In addition, the training algorithms based on generalized Fréchet distance and on orthogonal basis expansion are respectively adopted to train the RBF-PNN. The accuracy based on generalized Fréchet distance is superior to on orthogonal basis expansion. This is because the algorithm based on orthogonal basis expansion has fitting error and truncation error inevitably.

## 6. CONCLUSIONS

A RBF-PNN training method based on GA and SA hybrid optimization strategy is put forward in this paper. According to the network training objective function, by building a generalized Fréchet distance used to measure similarity between time-varying function samples, the functional optimization problem of RBF-PNN training is converted into extremal optimization solving of multivariate function, and the global optimization solution of the network parameters is obtained in the feasible solution space. It also has certain reference value for other machine learning and complex function optimization problems.

**Authors**

**Wang Bing**, born in 1982, is a Ph.D. candidate in Northeast Petroleum University. She received her Bachelor and Master degrees in Northeast Petroleum University, DaQing province China in 7/2004, 4/2007 respectively. From 2007 to now she is a lecturer at Department of Computer & Information Technology, Northeast Petroleum University. She majors in process neural networks, intelligence information processing and data mining etc. She has published more than 5 refereed journal and conference papers. 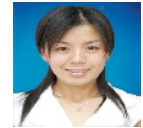

**Meng Yao-hua**, born in 1981, is a Ph.D. candidate in the Harbin Institute of Technology, Communication Research Center. He received his Bachelor and Master degrees in Northeast Petroleum University, DaQing province China in 7/2003, 8/2007 respectively. From 2007 to 2010 he was a lecturer at Department of Electrical and Electronic Engineering, Heilongjiang Bayi Agricultural University. During the year 2011, he was a visiting scholar at Harbin Institute of Technology. In August 2012, he joint Centre for massive dataset mining at Harbin Institute of Technology as a researcher. His research of interests includes process neural networks, signal and information processing etc. 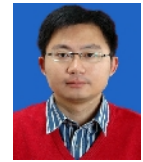

**Yu Xiao-hong**, born in 1982, is an engineer in Exploration and Development Research Institute of Daqing Oilfield Company, China. She received her Bachelor and Master degrees in Northeast Petroleum University, DaQing province China in 7/2005, 4/2008 respectively. From 2008 to now she works in Daqing Oilfield Company. Her interests include oil field data identification, artificial neural networks, intelligent information processing etc. 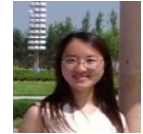